\PassOptionsToPackage{table,xcdraw}{xcolor}
\documentclass[sigconf,nonacm]{acmart}
\AtBeginDocument{%
  }

\usepackage{multirow}
\usepackage{algorithm}
\usepackage{algorithmic}
\usepackage{balance}

\copyrightyear{2026}
\acmYear{2026}
\setcopyright{cc}
\setcctype{by}
\acmConference[MM '26]{Proceedings of the 34th ACM International Conference on Multimedia}{November 10--14, 2026}{Rio de Janeiro, Brazil}
\acmBooktitle{Proceedings of the 34th ACM International Conference on Multimedia (MM '26), November 10--14, 2026, Rio de Janeiro, Brazil}
\acmDOI{10.1145/3767308.3834929}
\acmISBN{979-8-4007-2213-4/2026/11}

\begin{document}

\title{Single Point, Full Mask: Velocity-Guided Level Set Evolution for End-to-End Amodal Segmentation}

\author{Zhixuan Li}
\affiliation{%
  \department{College of Computing and Data Science}
  \institution{Nanyang Technological University}
  \city{Singapore}
  \country{Singapore}
}

\author{Yujia Liu}
\affiliation{%
  \department{School of Computer Science\\National Key Laboratory for Multimedia Information Processing\\National Engineering Research Center of Visual Technology}
  \institution{Peking University}
  \city{Beijing}
  \country{China}
}

\author{Chen Hui}
\affiliation{%
  \institution{Nanjing University of Information Science and Technology}
  \city{Nanjing}
  \country{China}
}

\author{Chenyue Song}
\affiliation{%
  \institution{Harbin Institute of Technology}
  \city{Harbin}
  \country{China}
}

\author{Weisi Lin}
\authornote{Corresponding author.}
\affiliation{%
  \department{College of Computing and Data Science}
  \institution{Nanyang Technological University}
  \city{Singapore}
  \country{Singapore}
}

\renewcommand{\shortauthors}{Zhixuan Li, Yujia Liu, Chen Hui, Chenyue Song, and Weisi Lin}

\begin{abstract}
Amodal segmentation aims to recover complete object shapes, including occluded regions, serving as an essential technique for user-centric multimedia authoring and object-level visual manipulation. Existing methods typically rely on informative prompts, such as bounding boxes or dense visible masks, which heavily degrade the user experience and interaction efficiency in real-world multimedia applications. While recent interactive paradigms (e.g., the Segment Anything Model) support lightweight point-based interactions, they often perform direct mask regression. Crucially, the opaque nature of these direct-regression models offers no visual explainability regarding how occluded structures are inferred, conflicting with the growing demand for interpretable multimedia systems. To address these limitations, we propose VELA, an end-to-end VElocity-driven Level-set Amodal segmentation method that enables explicit and transparent contour evolution driven by simple point clicks. VELA constructs an initial level set function from visual features and the user's point input, which then progressively evolves into the final amodal mask under the guidance of a shape-specific motion field predicted by a fully differentiable network. This mechanism learns to generate evolution dynamics at each step, ensuring that the spatial reasoning process is geometrically grounded, topologically flexible, and visually explainable to the user. Extensive experiments on COCOA-cls, D2SA, and KINS benchmarks demonstrate that VELA outperforms existing methods that use bounding-box or dense visible-mask prompts while requiring only a single-point prompt, validating the effectiveness of explainable geometric modeling for interactive multimedia tasks.
\end{abstract}


\maketitle

\section{Introduction}

The conventional visible segmentation task focuses solely on observable areas and is typically formulated as a pixel-wise classification problem, without inferring the shapes of occluded regions. In contrast, the amodal segmentation task aims to predict complete object masks, including both visible and occluded parts, by reasoning about the full object shape from the available visual cues. Despite its inherent challenges, amodal segmentation is rapidly becoming a cornerstone for human-centric multimedia applications, particularly in interactive image editing~\cite{sun2022ide,michel2023object}, object-level visual compositing~\cite{song2023objectstitch,qu2026deocc}, and the structural control of AI-generated content (AIGC)~\cite{wang2023imagen,manukyan2023hd}. In everyday multimedia creation, target subjects are frequently obscured by complex foreground elements, scene clutter, or dynamic human activities. Empowering users to easily recover the complete spatial extent of these occluded objects provides the essential geometric foundation required for tasks like seamless background replacement, object removal, and high-fidelity visual composition. Furthermore, facilitating this recovery through intuitive user interactions while ensuring the reasoning process remains visually transparent is a critical requirement for next-generation multimedia editing workflows. More broadly, robust and efficient multimedia perception has been studied under input perturbations, open-vocabulary recognition, domain shifts, and mobile deployment~\cite{liu2025norm,jin2024llms,jiang2023modify,zhang2023black,zhou2025care}.

\begin{figure*}[ht]
    \centering
    \includegraphics[width=\linewidth]{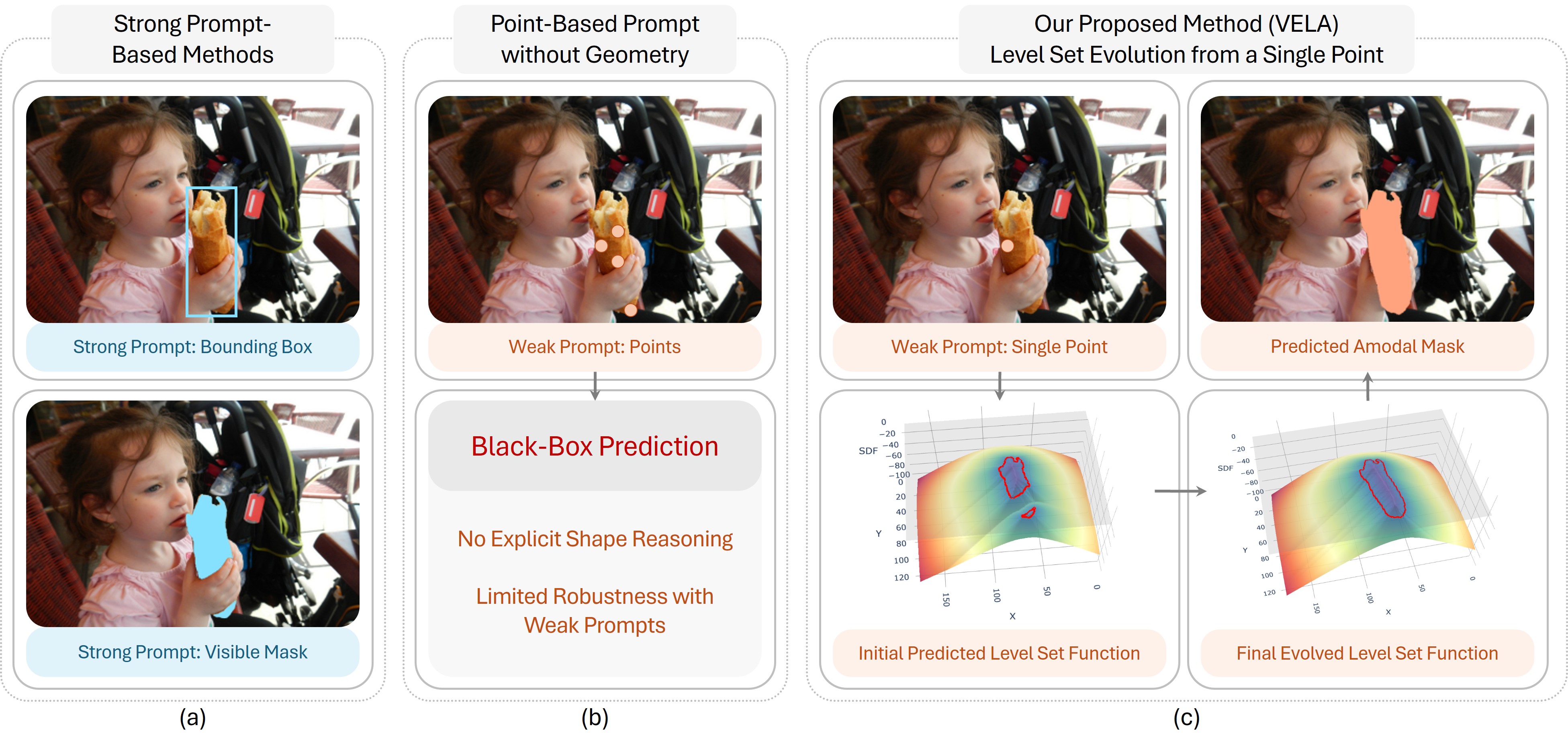}
    \caption{Comparison of amodal segmentation paradigms. (a) Conventional methods rely on bounding-box or dense visible-mask prompts.
    (b) Recent approaches support point prompts but directly regress masks without explicit shape reasoning, resulting in limited robustness under complex occlusions.
    (c) VELA introduces explicit shape evolution via Level Set, effectively exploiting a single-point prompt for interpretable amodal segmentation.}
    \Description{Three panels compare amodal segmentation paradigms: conventional methods use bounding-box or dense visible-mask prompts, direct-regression methods use point prompts without explicit shape reasoning, and VELA uses a single point to drive an explicit Level Set evolution toward the complete amodal mask.}
    \label{fig:introduction}
\end{figure*}

Existing state-of-the-art amodal segmentation methods~\cite{zhan2020self,ling2020variational,zheng2021visiting,nguyen2021weakly,liu2025towards} mostly rely on informative input prompts, such as dense visible masks~\cite{zhan2020self,ling2020variational,zheng2021visiting,nguyen2021weakly,gao2023coarse,zhan2024amodal} or bounding boxes~\cite{liu2024blade,liu2025towards}, to guide the prediction of amodal masks covering occluded regions. As shown in Fig.\ref{fig:introduction} (a), while these methods have achieved notable performance, obtaining such prompts is often impractical or expensive in real-world scenarios, significantly limiting the applicability of these approaches. Related efforts to reduce data and annotation demands include few-shot multi-view learning, prototype refinement, and monocular 3D perception based on masked autoencoding or weak supervision~\cite{zhou2023fewshot,zhou2022hierarchical,jiang2024monomae,jiang2024weakly}. 
Recent models like the Segment Anything Model (SAM)~\cite{kirillov2023segment,xiong2024efficientsam,ke2024segment} have introduced support for point-based prompts, offering a more lightweight form of guidance. However, as shown in Fig.\ref{fig:introduction} (b), these approaches typically predict masks directly from these sparse point-based prompts without an explicit mechanism to reason about the shape of occluded regions. As a result, their generalization ability remains limited in complex occlusion scenarios, where point prompts alone provide insufficient guidance for accurately inferring the complete amodal mask. 

Another key limitation of existing approaches~\cite{zhan2024amodal,tai2025segment,li2023oaformer,li2023muva,chen2023amodal} lies in their black-box nature, which lacks geometric interpretability in the amodal mask prediction process. Specifically, most existing methods adopt CNN-based~\cite{zhang2019learning,hu2019sail,qi2019amodal,follmann2019learning,zhan2020self,xiao2021amodal,li20222d,li2023gin} or Transformer-based~\cite{tran2022aisformer,li2023oaformer,tran2024shapeformer,li2025aura} architectures that treat amodal mask prediction as a direct regression task, without modeling the underlying morphological processes involved in shape segmentation. Consequently, these models offer little transparency into the internal reasoning behind their predictions, providing limited understanding of how occluded regions are inferred. This absence of explicit geometric reasoning not only limits the interpretability of the model’s predictions but also makes it difficult to analyze or improve model performance, especially in scenarios involving complex occlusions where understanding how the model reasons about fine-grained shapes is crucial for reliable prediction. Motivated by these limitations, we pose the following question: \textit{is it possible to achieve high-quality amodal segmentation using only a single-point prompt by fully leveraging a geometrically interpretable approach?}

To this end, we propose a novel approach named \textbf{VELA}, a \textbf{VE}locity-driven \textbf{L}evel-set \textbf{A}modal segmentation method, performing amodal segmentation via Level Set evolution with only a \textit{single point} as input prompt, as shown in Fig.\ref{fig:introduction} (c). Specifically, we begin by embedding the point prompt, which then interacts with the extracted image features to generate an initial level set function, representing the starting state of the object's amodal mask. This initial function is then progressively updated through a guided evolution process to expand toward the final high-quality amodal mask. 
To enable precise and adaptive guidance during this evolution, we design a fully differentiable network that learns to predict a shape-specific velocity field, which drives the contour evolution at each step. This velocity prediction network dynamically generates the up-to-date velocity field based on the current state of the evolving contour, providing context-aware guidance that continuously steers the contour toward the complete amodal shape effectively.

This evolution process simulates the gradual expansion of the object’s boundary with the step-by-step guidance generated based on contour status and learned prior knowledge to evolve toward the target shape. In contrast to conventional approaches that directly predict the final mask in a single step, our method models segmentation as a geometrically grounded and iterative process. This step-wise evolution not only enables more interpretable shape reasoning but also provides greater flexibility in handling complex occlusions and diverse object geometries. This is thanks to the Level Set paradigm’s inherent ability to explicitly represent and continuously evolve object contours in a geometrically interpretable and topologically flexible manner.
Building upon the unique characteristics of our Level Set-based evolution, VELA introduces a novel paradigm for amodal segmentation with single-point prompts. Moreover, the end-to-end framework allows adaptive modeling of diverse object shapes and occlusion patterns, alleviating limitations of prior black-box approaches with better transparency.

We extensively evaluate VELA on amodal segmentation benchmarks, including COCOA-cls, D2SA, and KINS. Results show that our Level Set-based approach with a single-point prompt outperforms methods using bounding-box or dense visible-mask prompts, demonstrating the effectiveness of geometry-aware modeling.

Our contributions are summarized as follows:
(1) We propose VELA, which, to the best of our knowledge, is the first amodal segmentation method that leverages Level Set evolution with single-point prompting. VELA achieves high-quality amodal segmentation with only a \textit{single point} as input, reducing reliance on bounding-box and dense visible-mask prompts.
(2) We introduce a fully differentiable framework that learns a shape-specific velocity field to explicitly guide the contour evolution process, enabling geometrically grounded and interpretable mask prediction.
(3) We demonstrate that our method performs well on amodal segmentation datasets using only a single-point prompt, showing its generalization capability in complex occlusion scenarios.

\section{Related Work}

\subsection{Amodal Segmentation}

Since the amodal segmentation task proposed by~\cite{li2016amodal}, various approaches have been developed from different perspectives. Several approaches~\cite{zhang2019learning,hu2019sail,qi2019amodal,xiao2021amodal,tran2022aisformer,li2023oaformer} are designed based on existing CNN-based or Transformer-based instance segmentation frameworks to tackle the amodal segmentation problem using only the input image. Meanwhile, many approaches~\cite{zhan2020self,ling2020variational,zheng2021visiting,nguyen2021weakly,gao2023coarse,liu2024blade} reformulate the task as predicting the amodal mask from the image and additional prompts, such as visible or amodal bounding boxes, or visible masks. Recent work further models shape diversity explicitly through a shape-aware sparse mixture-of-experts framework~\cite{li2025shapemoe}. 
However, bounding-box and dense visible-mask prompts for target objects are often impractical or costly to obtain in real-world scenarios. Such high demands on user input reduce the practicality of these methods and limit their ability to generalize to diverse scenarios, particularly in interactive applications. Beyond 2D amodal masks, recent studies extend scene understanding to multimodal 3D segmentation, intention-aware planning, view-robust action generation, camera-robust localization, and geometry-aware vision-language modeling~\cite{jiang2024multimodal,jiang2025exploring,li2026camvla,jiang2026camera,li2026vlm3d}.

To this end, in this paper, we investigate single-point prompts as guidance for amodal mask segmentation, thereby alleviating the dependence on bounding-box and dense visible-mask prompts. A natural solution is to leverage recent popular models, such as the Segment Anything Model (SAM), which are designed to accept point-based prompts as input. However, SAM focuses on visible mask segmentation with a direct prediction strategy and lacks explicit reasoning for occluded regions, which limits its ability to infer complete amodal masks, especially when only sparse point prompts are available that offer only coarse guidance of the visible parts of target objects. This motivates the need to explore how high-quality amodal segmentation can be achieved using only single-point prompts effectively.

\subsection{Level Set Methods for Segmentation}

Level set methods offer a powerful framework for image segmentation by implicitly encoding object boundaries as zero-level contours of higher-dimensional functions. Their capacity to represent arbitrary shapes and topological changes, such as merging and splitting, makes them particularly appealing for modeling complex object structures. Traditional approaches~\cite{caselles1997geodesic,kass1988snakes,malladi2002shape} typically focus on minimizing hand-crafted energy functions that capture low-level appearance cues, suffering from sensitivity to initialization and limited generalization ability.

To address these limitations, recent methods~\cite{kim2019mumford,homayounfar2020levelset,li2022box} have integrated the level set with deep learning frameworks. Some methods~\cite{ngo2013left,wang2019delse} improve initialization by using networks to predict coarse masks, edge probability maps, or contour priors that serve as starting level-set functions. Several methods utilize the level set from different aspects, including for post-processing for semi-supervised segmentation~\cite{tang2017deep}, integrating into the loss function for shape regulation~\cite{hu2017deep}, and denoising edge annotations~\cite{acuna2019devil}. 

Different from prior deep level set methods designed for visible-object segmentation, VELA targets amodal completion under a single-point prompt. It predicts a state-dependent velocity field at each iteration rather than relying on a static energy or contour map.

\begin{figure*}[ht]
    \centering
    \includegraphics[width=\linewidth]{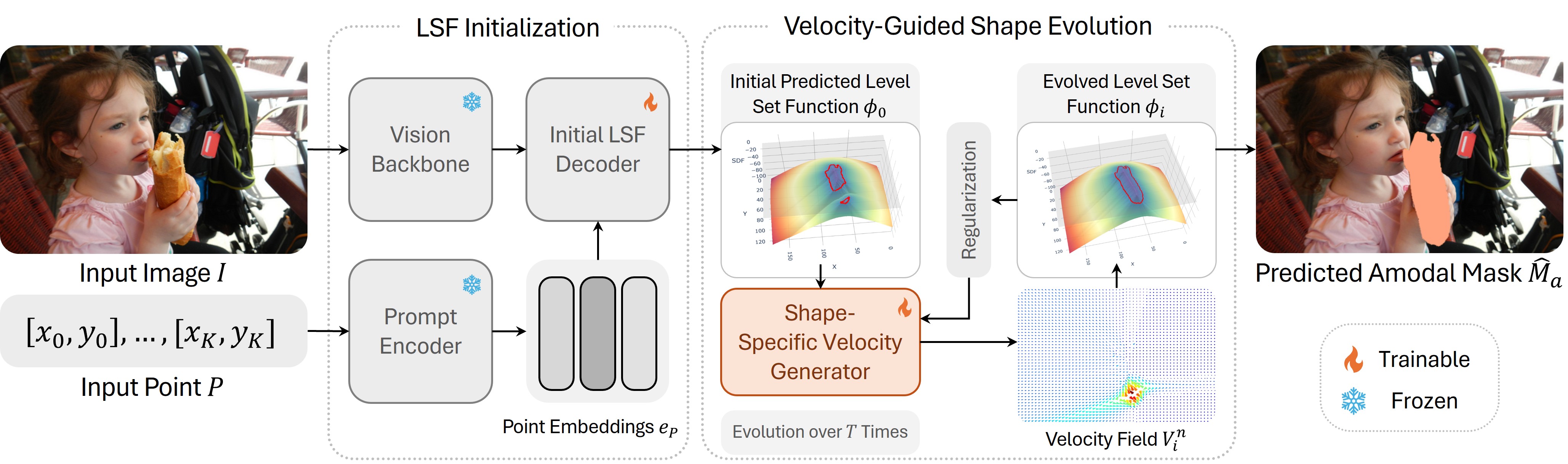}
    \caption{Overall architecture of the proposed VELA. (a) Level Set Function (LSF) Initialization: Given an input image $I$, the Vision Backbone extracts global image features, while the Prompt Encoder processes the point-level prompt $P$ to generate point embeddings $e_P$ that encode coarse localization of the target object. The Initial LSF Decoder then predicts the initial level set function $\phi_0$ using both the image features and point embeddings. (b) Velocity-Guided Shape Evolution: Starting from $\phi_0$, the shape evolves over $T$ steps. At each step $i$, the current level set function $\phi_i$ is fed into the Shape-Specific Velocity Generator to dynamically produce a velocity field $V^n_i$, which guides the contour’s deformation towards the final shape. The updated $\phi_{i+1}$ is then regularized and used for the next step. After $T$ steps, the final level set function $\phi_T$ is thresholded to obtain the predicted amodal mask $\hat{M}_a$. Best viewed in color.}
    \Description{A two-stage pipeline. The first stage combines image features and a point-prompt embedding to initialize a Level Set Function. The second stage repeatedly predicts a shape-specific velocity field, evolves and regularizes the contour for $T$ steps, and thresholds the final Level Set Function to obtain the amodal mask.}
    \label{fig:overall_architecture}
\end{figure*}

\section{Methodology}

We first define the amodal segmentation task and outline the VELA architecture. We then describe level set initialization, velocity-guided shape evolution, shape-specific velocity generation, and the end-to-end training objective.

\subsection{Overall Architecture}

Given an input image $I$ and an optional prompt $P$, the goal of amodal segmentation is to predict the complete amodal mask $\hat{M}_a$ of the target object. The prompt $P$ can take various forms, such as bounding boxes, masks, or points. In this work, we focus on point-based prompts, which offer coarse but efficient guidance for localizing the object to be segmented.

The proposed VELA consists of the following modules, as illustrated in Fig.~\ref{fig:overall_architecture}. 
First, we introduce feature extraction from inputs. Specifically, the backbone network processes the input image $I$ to extract the global feature representation $F$ of the entire scene. Simultaneously, the prompt encoder takes the point prompt $P$ as input and produces a sparse embedding $e_P$, which encodes coarse localization information about the target object to be segmented. A \textit{single point} prompt for the target object can provide sufficient guidance for effective amodal segmentation, while VELA also supports multiple point prompts if available.

Next, we introduce the evolution process guided by shape-specific velocity fields. A mask decoder first predicts the initial level set function $\phi_0$, which is then iteratively evolved over $T$ steps to produce the final level set function $\phi_T$. At each step $i \in \{0, \ldots, T-1\}$, $\phi_{i+1}$ is obtained by evolving $\phi_i$ under a dynamically generated, shape-specific velocity field. This adaptive guidance progressively deforms the zero-level contour toward the target amodal boundary. After $T$ evolution steps, the final level set function $\phi_T$ is thresholded at zero to produce the predicted binary amodal mask.

\begin{algorithm}[tb]
\caption{Velocity-Guided Level Set Evolution}
\label{alg:vela_evolution}
\begin{flushleft}
\textbf{Input:} Initial level set function $\phi_0$, number of steps $T$, time step $dt$ \\
\textbf{Components:} Velocity predictor $\mathcal{F}_v(\cdot)$, distance regularizer $\mathcal{R}(\cdot)$ \\
\textbf{Output:} Final evolved level set function $\phi_T$
\end{flushleft}
\begin{algorithmic}[1]
\STATE Initialize $\phi \leftarrow \phi_0$
\FOR{$i = 0$ to $T-1$}
    \STATE Predict normal velocity $V^n_i \leftarrow \mathcal{F}_v(\phi)$
    \STATE Update $\phi \leftarrow \phi - dt \cdot V^n_i$
    \STATE Apply regularization: $\phi \leftarrow \phi + \mu \cdot \mathcal{R}(\phi)$
\ENDFOR
\STATE \textbf{return} $\phi_T \leftarrow \phi$
\end{algorithmic}
\end{algorithm}

\subsection{Level Set Function Initialization}

The level set function is initialized by predicting $\phi_0$ using a mask decoder, which takes the global image feature $F$ and the sparse prompt embedding $e_P$ as inputs. For a 2D binary mask $M$ that contains a 2D curve $\mathcal{C}$ of the target object, the level set function $\phi$ provides a higher-dimensional implicit representation of $M$. Specifically, $\phi$ is a 3D surface whose zero level set corresponds to the object boundary curve $\mathcal{C}$ in the 2D mask $M$. Here, we define background regions as negative and foreground regions as positive. Accordingly, the binary mask $M$ can be obtained by thresholding the level set function $\phi$ at zero, retaining values greater than zero as the foreground. We can formulate the aforementioned correspondence as follows: 

\begin{equation}
M(x, y) =
\begin{cases}
1, & \text{if } \phi(x, y) > 0 \\
0, & \text{otherwise}
\end{cases}
\quad \text{with } \mathcal{C} = \partial M
\end{equation}

Here, $x$ and $y$ represent the 2D spatial coordinates of the image, and $\partial$ denotes the boundary operator. Accordingly, $\mathcal{C} = \partial M$ defines the object contour corresponding to the boundary of the binary mask $M$.

In our design, the predicted initial level set function $\phi_0$ corresponds to the mask of visible regions for the target object, since the input point prompt only provides very coarse guidance to the visible part of the object. Compared to the traditional level set approach, our initialization is predicted by the network in an end-to-end approach.

\subsection{Velocity-Guided Shape Evolution}

The proposed VELA is designed to fully leverage the information provided by the sparse prompt to achieve high-quality amodal segmentation through a geometrically interpretable approach. A velocity-guided shape evolution process is introduced to drive the morphological transformation of the level set function from the initial state $\phi_0$, corresponding to the predicted visible mask, toward the complete object shape represented by the ground-truth amodal mask $M_a$. After $T$ updates, the final predicted state is $\phi_T$. The complete evolution procedure is outlined in Alg.~\ref{alg:vela_evolution}.

Specifically, the initial level set function $\phi_0$ is evolved over $T$ steps to expand the initial contour of the visible mask toward the target amodal mask, which typically encompasses a larger region than the visible mask. Based on the level set theory, for step $i \in \{0, ..., T-1\}$, the progressively updating process can be formulated as:

\begin{equation}
    \phi_{i+1} = \phi_{i} + dt \frac{\partial \phi}{\partial t},
\end{equation}

\noindent where $dt$ denotes the time step, representing the time amount for the contour to move per step, and $\frac{\partial \phi}{\partial t}$ represents the temporal derivative of the level set function, governing its deformation over time. In classical level set theory, the contour evolves along its normal direction, and the update term is defined as:

\begin{equation}
\frac{\partial \phi}{\partial t}
= - V^n_i \lVert \nabla \phi \rVert,
\label{eqn:ori_update_term}
\end{equation}

\noindent where $V^n_i$ denotes the predicted normal velocity field at step $i$, which controls the speed of evolution along the normal direction, and $\lVert \nabla \phi \rVert$ is the gradient magnitude that enforces evolution along the normal direction.

To simplify the computation and ensure numerical stability, we apply a regularization term after each evolution step to maintain the level set function $\phi_i$ as a signed distance function (SDF), which keeps $\lVert \nabla \phi \rVert$ close to 1 throughout the evolution. Under this condition, the evolution equation can be simplified by treating the gradient norm as nearly constant and close to 1. As a result, Eqn.~\ref{eqn:ori_update_term} becomes:

\begin{equation}
\frac{\partial \phi}{\partial t} = - V^n_i,
\end{equation}

\noindent leading to a simplified update rule:

\begin{equation}
\phi_{i+1} = \phi_i - dt \cdot V^n_i.
\end{equation}

\noindent This formulation allows the evolving contour to be directly guided by the predicted scalar velocity field $V^n_i$, while the distance regularization ensures that the geometric meaning of the level set function is preserved throughout the evolution process. 

To implement this, we adopt a distance regularization term $\mathcal{R}(\phi)$ inspired by the double-well potential in DRLSE~\cite{li2010distance}, which encourages $\|\nabla \phi\|$ to stay close to 1. Specifically, at each step, we add a regularization update:

\begin{equation}
\widetilde{\phi}_{i+1} = \phi_i - dt \cdot V^n_i, \qquad
    \phi_{i+1} = \widetilde{\phi}_{i+1} + \mu \cdot \mathcal{R}(\widetilde{\phi}_{i+1}),
\end{equation}

\noindent where $\widetilde{\phi}_{i+1}$ is the intermediate state after the velocity update and $\mu$ is a weighting factor. As in Alg.~\ref{alg:vela_evolution}, regularization is applied to this intermediate state. This regularization supports maintaining the signed distance property of $\phi$, thereby enabling the approximation that the gradient norm remains close to 1 throughout the evolution.

Finally, the amodal segmentation mask is obtained by applying a zero threshold to the final level set function $\phi_T$, with the predicted foreground amodal mask $\hat{M}_a$ defined as $\{(x, y) \mid \phi_T(x, y) > 0\}$.

\subsection{Shape-Specific Velocity Generation}
To enable accurate and interpretable evolution of the level set function, we introduce the Shape-Specific Velocity Generator that dynamically predicts the normal velocity field $V^n_i$ at each evolution step. Specifically, we employ a lightweight UNet architecture that takes the current level set function $\phi_i$ as input and outputs a scalar velocity map that guides the contour deformation. This predicted velocity field encodes the directional cues required to deform the current visible contour toward the final amodal shape, allowing the model to learn shape-aware motion patterns conditioned on the evolving mask geometry dynamically.

\paragraph{Network Architecture of the Shape-Specific Velocity Generator}
To guide the contour evolution in VELA, we employ a UNet network to dynamically predict the velocity field at each iteration. The input to this module is the current level set function $\phi_i$, and the output is a single-channel velocity map $V_i$ that encodes the scalar speed of contour evolution at each spatial location. 

Internally, we adopt a lightweight UNet architecture to extract spatial features from the input $\phi_i$ and generate the velocity field, enabling flexible and data-adaptive evolution dynamics. By predicting the velocity field in a differentiable manner, this module allows the model to learn how to adjust the evolution process based on intermediate contour states, which is essential for accurate amodal mask recovery.

\paragraph{Gradient Detachment of Initial Level Set}
We detach the initial level set function $\phi_0$ before evolution to prevent the amodal supervision applied to later states from directly altering the visible-region initializer. This separates initial localization from subsequent amodal shape evolution.

\subsection{Training Objective}

To supervise the evolution of the level set function toward the desired amodal shape, we define a loss that provides step-wise guidance during training. After each evolution step $i \in \{0, \ldots, T-1\}$, we apply supervision to the updated level set function $\phi_{i+1}$. The loss therefore covers all evolved states $\phi_1, \ldots, \phi_T$, excluding the initial state $\phi_0$. We apply a smooth approximation of the Heaviside function $H(\phi_{i+1})$ to obtain a differentiable soft mask, which maps positive values to foreground and negative values to background, enabling end-to-end learning. We then compute a binary cross-entropy (BCE) loss between the soft prediction and the ground-truth amodal mask $M_a$:

\begin{equation}
\mathcal{L}_{\text{evo}} = \sum_{i=0}^{T-1} \text{BCE}(H(\phi_{i+1}), M_a),
\end{equation}

This cumulative loss encourages the model to learn meaningful velocity fields that progressively deform the evolving contour toward the complete amodal mask.

\section{Experiments}

To evaluate the effectiveness of the proposed VELA, we conduct experiments on three challenging datasets. Compared with existing methods, VELA achieves state-of-the-art performance. We illustrate the comparison both quantitatively and qualitatively. Finally, extensive ablation studies are conducted to validate the design of VELA. 

\begin{table*}[t]
\centering
\setlength\tabcolsep{6pt}
\caption{
    Comparison with state-of-the-art approaches on the COCOA-cls dataset (validation split), the D2SA dataset (validation split), and the KINS dataset (testing split). Metrics mIoU$_{\text{full}}$ and mIoU$_{\text{occ}}$ are employed and higher values indicate better performance. Zero-shot methods are evaluated without training on the target datasets. Pix2Gestalt~\cite{ozguroglu2024pix2gestalt} uses a synthetic amodal training dataset containing 873K images. 
    For the prompt type, ``Mask'' denotes the dense visible mask of the target object, while ``Point'' denotes a single sparse point located within the visible region.
}
\begin{tabular}{lcccccccc}
\toprule
\multicolumn{1}{l}{\multirow{2}{*}{Methods}} &
  \multicolumn{1}{c}{\multirow{2}{*}{Venue}} &
  \multicolumn{1}{c}{\multirow{2}{*}{Prompt}} &
  \multicolumn{2}{c}{COCOA-cls   (Val)} &
  \multicolumn{2}{c}{D2SA   (Val)} &
  \multicolumn{2}{c}{KINS   (Test)} \\ \cmidrule{4-9}
  \multicolumn{1}{c}{} &
  \multicolumn{1}{c}{} &
  \multicolumn{1}{c}{} &
  mIoU$_{\text{full}}\uparrow$ &
  mIoU$_{\text{occ}}\uparrow$ &
  mIoU$_{\text{full}}\uparrow$ &
  mIoU$_{\text{occ}}\uparrow$ &
  mIoU$_{\text{full}}\uparrow$ &
  mIoU$_{\text{occ}}\uparrow$ \\
\hline
\textit{Zero-shot Methods}  & & & & & & & & \\
\hline
SAM~\cite{kirillov2023segment}                & ICCV 2023 & Mask  & 73.10 &   -    & 84.65 &   -  & 75.88 &   -    \\
SDXL-Inpainting~\cite{podell2023sdxl}    & ICLR 2024 & Mask  & 73.65 &   -    & 80.53 &   -    & 76.19 &   -    \\
Pix2Gestalt~\cite{ozguroglu2024pix2gestalt}        & CVPR 2024 & Mask  & 79.08 &  -     & 81.82 &   -    & 81.45 &   -   \\
\hline
\textit{Fully-Supervised Methods}  & & & & & & & & \\
\hline
BCNet~\cite{ke2021deep}              & CVPR 2021 & -         & 15.10 &    -   & 74.90 &   -    & 44.00 &   -    \\
SLN~\cite{zhang2019learning}                & MM 2019   & -         & 31.50 &    -   & 31.20 &   -    & 14.20 &   -    \\
ORCNN~\cite{follmann2019learning}              & WACV 2019 & -         & 57.60 &    -   & 74.10 &   -    & 55.10 &   -    \\
Mask-RCNN~\cite{he2017mask}          & ICCV 2017 & -         & 63.85 &    -   & 74.62 &   -    & 60.13 &   -    \\
A3D~\cite{li20222d}                & ECCV 2022 & -         & 64.27 &    -   & 74.71 &   -    & 61.43 &   -    \\
GIN~\cite{li2023gin}                & TMM 2023  & -         & 72.46 &    -   & 78.23 &   -    & 68.31 &   -    \\
AISFormer~\cite{tran2022aisformer}          & BMVC 2022 & -         & 72.69 & 13.75 & 86.81 & 30.01 & 81.53 & 48.54 \\
PCNet~\cite{zhan2020self}              & CVPR 2020 & Mask  & 76.91 & 20.34 & 80.45 & 28.56 & 78.02 & 38.14 \\
VRSP~\cite{xiao2021amodal}               & AAAI 2021 & -         & 78.98 & 22.92 & 88.08 & 35.17 & 80.70 & 47.33 \\
SDAmodal~\cite{zhan2024amodal}           & CVPR 2024 & Mask  & 80.01 &   -    &   -    &  -     &   -    &  -     \\
C2F-Seg~\cite{gao2023coarse}            & ICCV 2023 & Mask  & 80.28 & 27.71  & 89.10 &   -    & 82.22 & 53.60  \\ 
\rowcolor[HTML]{EFEFEF} 
\textbf{VELA (Ours)}               & ACM MM 2026         & Point &   80.68    &  25.00     &   89.96    &  39.41     &  83.06     &   50.29   \\
\bottomrule
\end{tabular}
\label{tab:comparison_with_sota}
\end{table*}

\subsection{Datasets and Evaluation Metrics}

\paragraph{Datasets.} 
We conduct experiments on three publicly available datasets to evaluate the performance of VELA. Specifically, first, the COCOA-cls~\cite{follmann2019learning} dataset contains 3,501 images and 10,592 ground-truth masks across 80 categories, which is constructed based on the famous COCO~\cite{lin2014microsoft} dataset. The COCOA-cls dataset serves as a crucial benchmark due to its diverse and representative daily life scenarios. 
Second, the D2SA~\cite{follmann2019learning} dataset is designed for the shopping scenario, where self-checkout systems are commonly used. With 5,600 images and 28,720 ground-truth masks in 60 categories provided, the D2SA dataset can effectively examine the amodal segmentation ability of methods with its intrinsic occlusions constructed by overlapping merchandise.
Finally, built upon the KITTI~\cite{geiger2012we} dataset, the KINS~\cite{qi2019amodal} dataset comprises 7,474 and 7,517 images in the training and testing splits, respectively. 
It primarily captures street scenes, making it a valuable benchmark for assessing amodal segmentation performance in complex urban environments. 

For each dataset, fully supervised methods are trained on its official training split. Zero-shot methods are evaluated without training on the target datasets. All methods are evaluated on the same official validation or testing split. 
All experiments of VELA are conducted three times, and the average performance is reported to mitigate the impact of randomness.

\paragraph{Evaluation Metrics.}
Following~\cite{gao2023coarse}, we adopt mean Intersection over Union (mIoU) as the evaluation metric, computed over two regions: the entire amodal mask and the occluded regions. These are denoted as mIoU$_{\text{full}}$ and mIoU$_{\text{occ}}$, respectively. Larger values indicate better performance. 

\subsection{Implementation Details}

We implement our proposed VELA framework based on PyTorch~\cite{paszke2019pytorch}. VELA takes SAM 2~\cite{ravi2024sam} as the base structure, whose pretrained weights are used for initializing VELA.
The optimizer for gradient descent is AdamW. For computing infrastructure, all experiments are conducted on a machine equipped with an AMD 9950X3D with 96 GB system memory, an NVIDIA 5090 GPU with 32 GB of memory, and the Windows 11 operating system. There are no data augmentation techniques employed in the experiments of VELA. 

\begin{figure*}[ht]
    \centering
    \includegraphics[width=0.999\linewidth]{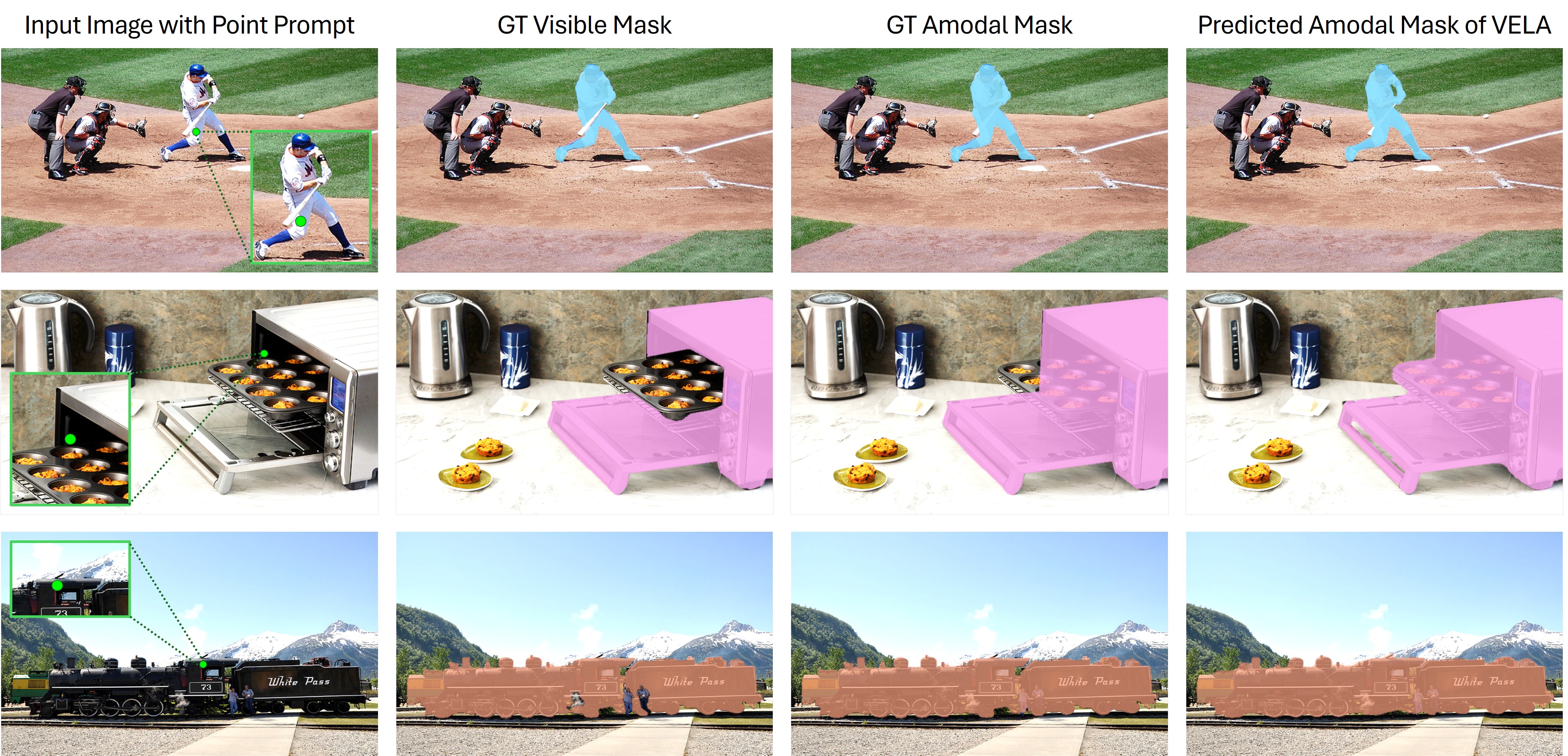}
    \caption{Visualization results of VELA. The green dot denotes the input point prompt. Best viewed in color.}
    \Description{Representative COCOA-cls examples showing VELA predictions for objects with complex occlusions and diverse shapes. In each example, a green dot marks the single point prompt used to guide recovery of the complete amodal mask.}
    \label{fig:qualitative_results}
\end{figure*}

\subsection{Comparison with State-of-the-Art Methods}

We compare the proposed VELA with state-of-the-art approaches, including both zero-shot and fully supervised methods. 
Specifically, the zero-shot baselines include SAM, SDXL-Inpainting, and Pix2Gestalt, which are evaluated without training on the target datasets. Pix2Gestalt is trained on a synthetic amodal segmentation dataset~\cite{ozguroglu2024pix2gestalt} comprising approximately 873,000 images with simulated occlusions and corresponding ground-truth amodal masks.
Fully-supervised methods, including VELA, are trained exclusively on the training splits of the three datasets, without utilizing any additional amodal segmentation data. Several methods use the visible mask of the target object as the input prompt, whereas VELA relies solely on a single point located within the visible region.

As shown in Tab.~\ref{tab:comparison_with_sota}, VELA achieves the highest mIoU$_{\text{full}}$ on all three benchmarks. For mIoU$_{\text{occ}}$, VELA exceeds all baselines with reported values except C2F-Seg on COCOA-cls and KINS; C2F-Seg uses a dense visible-mask prompt, whereas VELA uses a single sparse point.
Compared to the zero-shot method Pix2Gestalt, which is pretrained on a large-scale amodal segmentation dataset, VELA achieves gains of 1.60, 8.14, and 1.61 percentage points in mIoU$_{\text{full}}$ across the three benchmarks. VELA also surpasses the fully supervised C2F-Seg in mIoU$_{\text{full}}$ on all three datasets.
Despite using only a single point as the prompt, VELA therefore achieves higher full-mask accuracy than all compared methods that rely on visible-mask prompts. This result demonstrates the effectiveness of our geometrically interpretable Level Set evolution, which progressively transforms the visible region into the complete amodal shape under learned velocity guidance.

\begin{figure*}[htbp]
    \centering
    \includegraphics[width=0.999\linewidth]{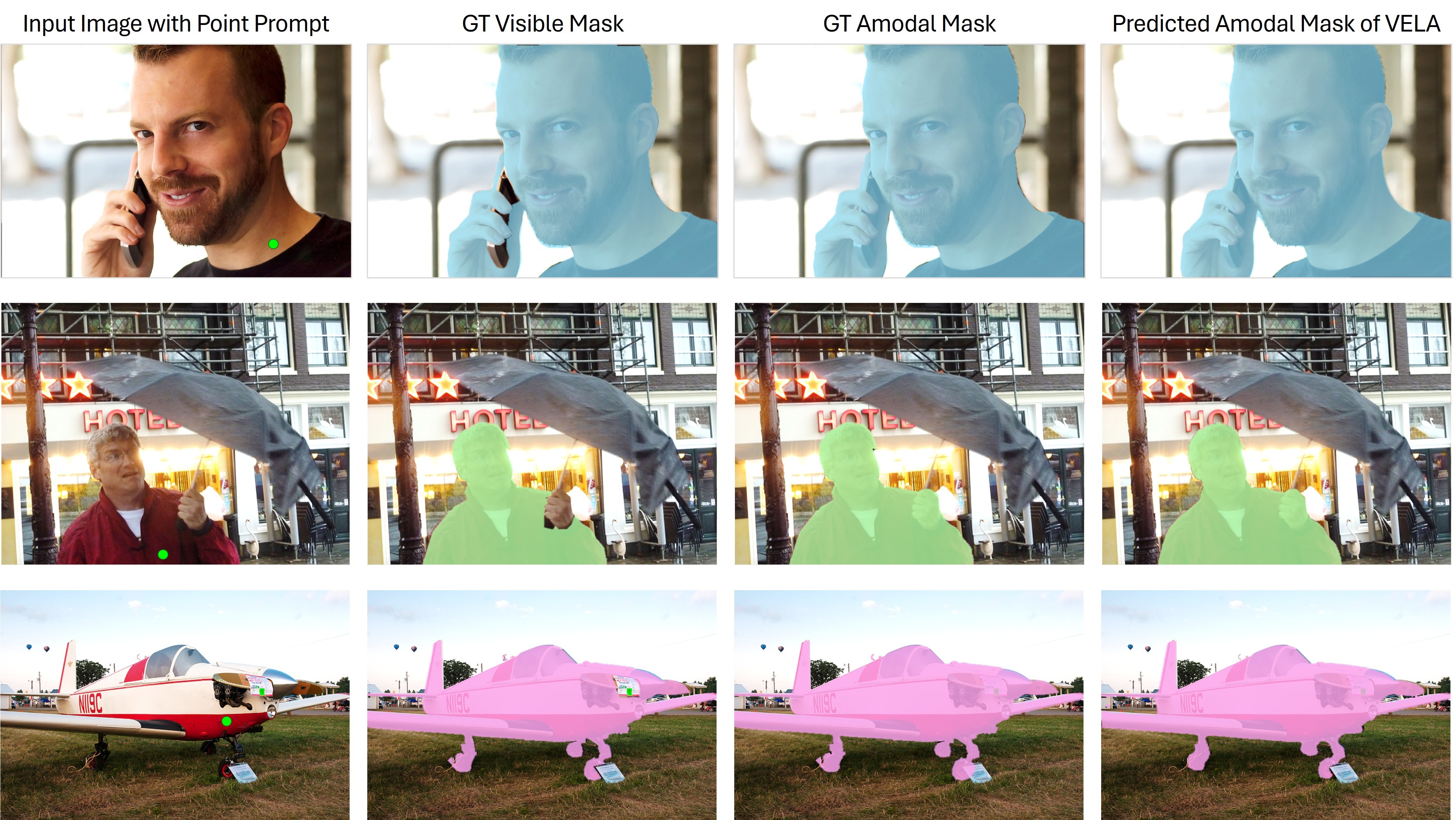}
    \caption{Visualization results of VELA. The green dot denotes the input point prompt. Best viewed in color.}
    \Description{Additional COCOA-cls examples showing VELA predictions under complex occlusions and varied object geometry. A green dot denotes the single input point from which VELA evolves the complete amodal mask.}
    \label{fig:supp_vis2}
\end{figure*}

Moreover, Fig.~\ref{fig:qualitative_results} and~\ref{fig:supp_vis2} show qualitative results of VELA on the COCOA-cls dataset. The qualitative examples show that VELA can recover coherent amodal shapes across diverse occlusion patterns, although highly irregular local boundaries remain challenging when visual evidence is limited.

\begin{table}[t]
\centering
\setlength\tabcolsep{4pt}
\caption{Ablation study of the evolution steps.}
\begin{tabular}{ccccc}
\toprule
Index & Evolution Steps $T$ & COCOA-cls      & D2SA           & KINS           \\
\hline
1     & 1                & 58.42          & 62.36          & 56.49          \\
2     & 2                & 69.67          & 79.20          & 66.99          \\
\rowcolor[HTML]{EFEFEF} 
3     & 3                & \textbf{80.68} & \textbf{89.96} & \textbf{83.06} \\
4     & 4                & 78.68          & 83.52          & \underline{80.58}    \\
5     & 5                & 79.09          & 88.75          & 77.66          \\
6     & 10               & \underline{79.75}    & \underline{89.65}    & 78.39       \\
\bottomrule
\end{tabular}
\label{tab:ab_evolution_steps}
\end{table}

\subsection{Ablation Study}

We conduct ablation studies on VELA to evaluate the effectiveness of its key components, including level set evolution steps, the number of input points, and the segmentation supervision strategy. Unless otherwise stated, all experiments are conducted using three evolution steps and a single-point prompt. Results are reported using the mIoU$_\text{full}$ metric on the COCOA-cls, D2SA, and KINS datasets.

\begin{table}[t]
\centering
\setlength\tabcolsep{5pt}
\caption{Ablation study of the number of input points.}
\begin{tabular}{ccccc}
\toprule
Index & \# Input Point(s) & COCOA-cls & D2SA  & KINS  \\ 
\hline
1     & 1              & 80.68     & 89.96 & 83.06 \\
2     & 5              & 80.72     & 90.12 & 83.51 \\
3     & 10             & 80.79     & 90.23 & 83.94 \\
4     & 20             & 80.83     & 90.51 & 83.96 \\
5     & 50             & 79.86     & 89.11 & 83.02 \\ \bottomrule
\end{tabular}
\label{tab:ab_input_points}
\end{table}

\begin{table}[t]
\centering
\setlength\tabcolsep{2pt}
\caption{Ablation study of the supervision design.}
\begin{tabular}{ccccc}
\toprule
Index & Supervision Design & COCOA-cls & D2SA  & KINS  \\ 
\hline
1     & Only final prediction              & 79.12     & 88.64 & 81.47 \\
2     & All evolved predictions              & \textbf{80.68}     & \textbf{89.96} & \textbf{83.06} \\ \bottomrule
\end{tabular}
\label{tab:ab_supervision_design}
\end{table}

\paragraph{Effect of Evolution Steps $T$.} 
As shown in Tab.~\ref{tab:ab_evolution_steps}, we evaluate the impact of varying the number of level set evolution steps $T$ in VELA. The best performance is observed when $T = 3$. While fewer steps may limit occlusion recovery, a large number of steps can introduce mild contour smoothing or deformation, which may have a certain impact on the quality of predicted amodal masks.

\paragraph{Effect of Input Points Number.}

As shown in Tab.~\ref{tab:ab_input_points}, we investigate the impact of using different numbers of point prompts. Increasing the number of input points brings modest improvements up to 20 points, but further increases offer little benefit and may introduce noise that affects performance. Notably, VELA performs well even with a single-point prompt, demonstrating its ability to effectively utilize sparse guidance for high-quality amodal segmentation.

\paragraph{Effect of Supervision Design.}

We investigate two supervision strategies by either applying the loss only to the final evolved prediction $\phi_T$ or supervising all updated states $\phi_1, \ldots, \phi_T$, including the final prediction and excluding the initial state $\phi_0$. As shown in Tab.~\ref{tab:ab_supervision_design}, incorporating supervision at all steps consistently improves performance across datasets. This design enables step-wise supervision for the velocity field, allowing the model to refine the shape evolution more effectively and maintain stability throughout the process.

\section{Discussions}

Level sets provide geometry-aware contour evolution and accommodate topology changes under occlusion. VELA combines semantic completion from learned velocity fields with contour continuity from regularization, although weak local cues can lead to oversmoothed fine boundaries. A single point minimizes interaction while supporting accurate amodal recovery; multiple points can provide more precise localization when available. Future extensions may incorporate upstream detections, temporal consistency, or language conditioning. Complementary multimedia components include adaptive acquisition and reconstruction~\cite{hui2025image,hui2023rate,hui2024s,hui2022multichannel}, while spatio-temporal modeling, watermarking, and source-camera identification offer further system-level capabilities~\cite{li2026stsmixer,hui2022spatiotemporal,hui2021adaptive,hui2022source}.

\section{Conclusion and Limitation Analysis}
VELA formulates single-point amodal segmentation as explicit contour evolution driven by a differentiable velocity generator. Unlike direct mask regression, VELA initializes a level set function from a single-point prompt and progressively evolves its zero-level contour using shape-specific velocity fields. This formulation combines learned semantic completion with geometry-aware contour evolution and exposes the intermediate deformation process before the final mask is obtained. Experiments on COCOA-cls, D2SA, and KINS demonstrate strong performance with sparse guidance, showing that a single point can provide an effective initialization for recovering complete object shapes under occlusion. Future work will investigate instance-adaptive stopping criteria and closed-loop interaction, allowing additional corrective prompts to refine uncertain contours while preserving the efficiency of sparse guidance.

\begin{acks}
This research was supported by the Ministry of Education of Singapore Tier3 grant MOET32022-0006. This research was partially funded by National Natural Science Foundation of China (62506009) and China Postdoctoral Science Foundation (2026M791599). This work was supported by the Natural Science Foundation of Jiangsu under Grant BK20250731, the General Program of China Postdoctoral Science Foundation under grant 2025M784450, the Startup Foundation for Introducing Talent of Nanjing University of Information Science and Technology under grant 2025r029.
\end{acks}

\clearpage

\bibliographystyle{ACM-Reference-Format}
\balance
\bibliography{sample-base}

\end{document}